\documentclass[11pt]{article}

\usepackage{acl}
\usepackage{xurl}
\usepackage{multirow}
\usepackage{times}
\usepackage{latexsym}
\usepackage{subcaption}

\usepackage[T1]{fontenc}

\usepackage[utf8]{inputenc}

\usepackage{microtype}

\usepackage{inconsolata}

\usepackage{graphicx}
\usepackage{booktabs}
\usepackage{dblfloatfix}
\usepackage{mdframed}
\usepackage{xcolor}
\usepackage{tcolorbox}
\tcbset{rqbox/.style={
    colback=gray!12,
    colframe=gray!40,
    boxrule=0.6pt,
    arc=3pt,
    left=6pt, right=6pt, top=6pt, bottom=6pt
}}

\title{One prompt is not enough: Instruction Sensitivity Undermines Embedding Model Evaluation}

\author{Yevhen Kostiuk$^*$ \\
  Aarhus University \\
  \texttt{ykost@cas.au.dk} \\\And
  Kenneth Enevoldsen$^*$ \\
  Aarhus University \\
  \texttt{kenneth.enevoldsen@cas.au.dk} \\}

\begin{document}
\maketitle
\begin{abstract}
{\let\thefootnote\relax\footnotetext{$^*$Equal contribution.}}

Instruction embedding models have become common among state-of-the-art models, however are evaluated using a single prompt per task. The single-point evaluation ignores a main problem of the instruction-based approach namely: sensitivity to the phrasing of the instruction. We present an empirical study of prompt sensitivity across 6 embedding models and 11 datasets. We show that reported scores misrepresent the distribution of scores over plausible prompts. The default prompt can both systematically understate or overstate performance. Furthermore, we show that the leaderboard ranking is not robust to prompt selection: a developer improve their rank by favorably selecting prompts, and under adversarial prompt selection any model can be promoted to first place. Our findings suggest that single-prompt evaluation is insufficient for instruction-tuned embedding models and that benchmarks should incorporate prompt robustness, either by evaluating over multiple prompts or by reporting sensitivity alongside point estimates.

The code can be found in our GitHub repository\footnote{{\footnotesize\sloppy\url{https://github.com/centre-for-humanities-computing/one-prompt-is-not-enough}}}.
\end{abstract}

\section{Introduction and Background}

Embeddings are foundational to modern NLP applications, serving as the backbone for tasks such as Retrieval-Augmented Generation (RAG), semantic search, zero-shot clustering as well as newer tasks such as agentic memory \cite{lewis2020retrieval, xu2026mem}. Recently, instruction-tuned embedding models initially popularized by \citet{su2023embeddertaskinstructionfinetunedtext} have become common among top-performing embedding models such as Qwen or the e5-series \cite{qwen3embedding, wang2024multilingual, enevoldsen2025mmtebmassivemultilingualtext}. Instructions (or prompts) enables models to capture task-specific nuances of the text, offering higher flexibility for various domains and objectives. However, while instructions allow for a high degree of user-flexibility they also allow instructions to be tuned to obtain better benchmark performance, thus posing a fundamental challenge of how to best measure, report and compare such models that do not oversell nor undersell their capabilities.

\begin{figure}
    \centering
    \includegraphics[width=\linewidth]{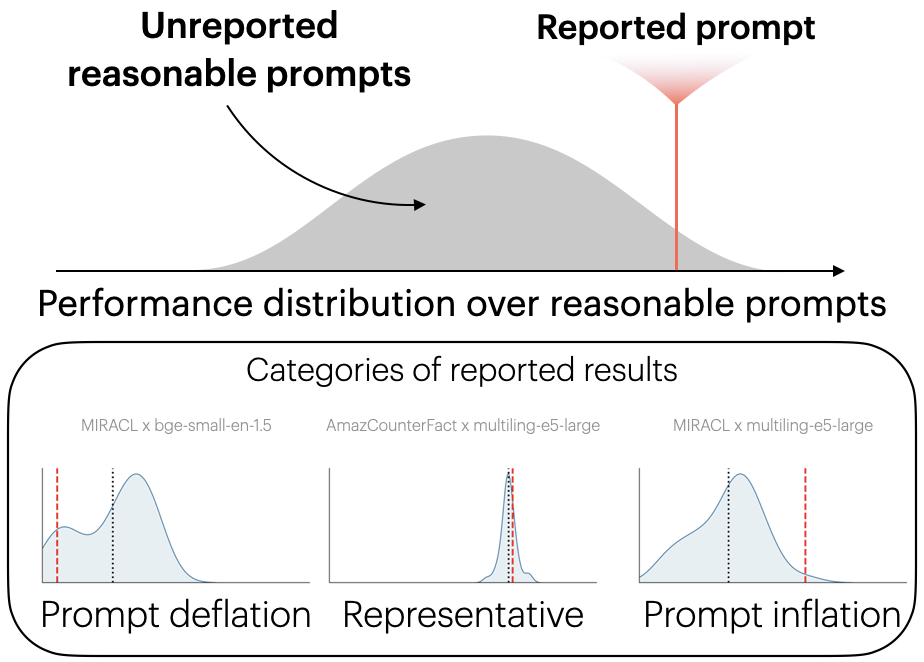}
    \caption{\textbf{Overview of our approach}, showing the currently unreported distribution of performances over reasonable prompts against the point estimate that is the reported results. Below that we show three possible option, either the reported result is representative of the distribution, it is lower than the (prompt deflation) or higher than (prompt inflation).}
    \label{fig:abstract}
\end{figure}

\begin{figure*}
    \centering
    \includegraphics[width=1.0\linewidth]{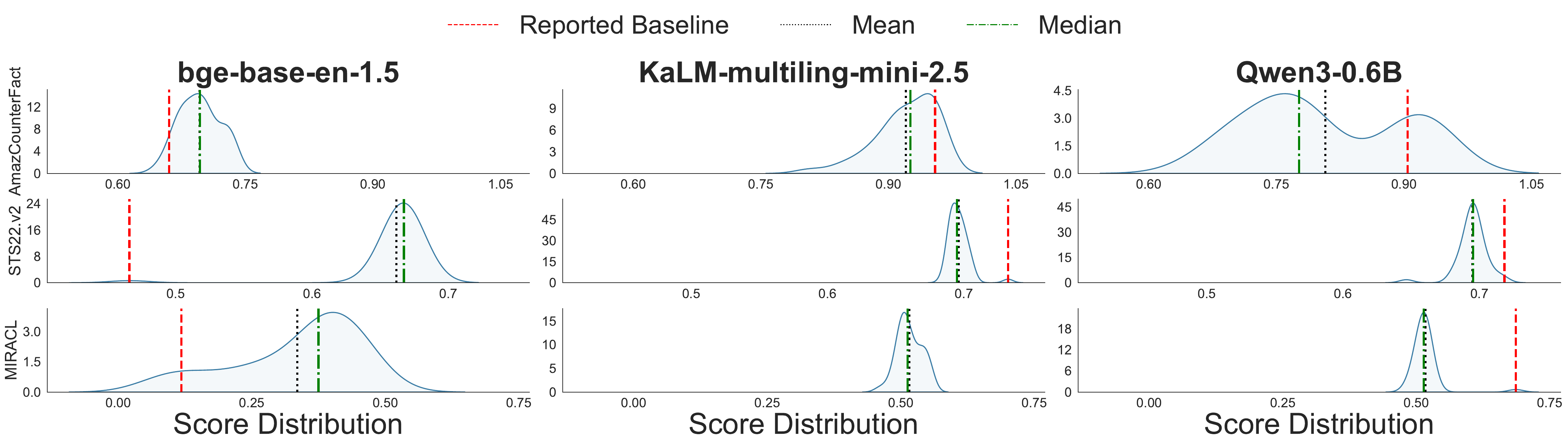}
    \caption{\textbf{Example of prompt deflation and inflation} on a sample of evaluated tasks and models. For all the scores and distributions see \autoref{sec:det_scores}. For KaLM model  on MIRACL task, mteb does not have reported score.}
    \label{fig:prompt_infl_defl_example}
\end{figure*}

The most common benchmarks for evaluating the quality of embedding are implemented using the \texttt{beir} for information retrieval \cite{thakur2021beir, kamalloo:2024} and multitask \texttt{mteb} frameworks, which includes popular benchmarks such as MTEB, MMTEB \cite{muennighoff2022mteb, enevoldsen2025mmtebmassivemultilingualtext}. \texttt{mteb} evaluate instruction-based models using only a single prompt either supplied by the task or the model contributor. This approach ignores a defining characteristic of these models: that they are sensitive to phrasing of the instruction. Relying on a single prompt risks misrepresenting actual model performance. Furthermore, it is possible to overwrite the prompt within the models implementation -- to allow customization to a specific prompt scheme -- making it possible for the model developers to selectively choose the best performing prompts for a given task. 

Current research on prompt sensitivity is focused mostly on generative LLMs \cite{wahle-etal-2024-paraphrase, leidinger2023languagepromptinglinguisticproperties, romanou2026brittlebenchquantifyingllmrobustness, mizrahi-etal-2024-state}, while instruction-based embedding models remains largely underexplored. While recent efforts have addressed overfitting via paraphrases \cite{frank-afli-2026-pteb}, no work systematically examines the effect of prompt variation on embedding evaluation. Our work bridges this gap by isolating the prompt as the primary variable to quantify its impact on reported performance.

In this paper, we question the validity of single-prompt evaluation for instruction-tuned embedding models. We investigate the following research questions: 

\begin{tcolorbox}[rqbox, colframe=gray!12]
\noindent (\textbf{RQ1}) To what extent are instruction-tuned embedding models sensitive to variations of prompts relevant to the task?\\
(\textbf{RQ2}) How do these variations influence the reported benchmark rankings?\\
(\textbf{RQ3}) To what degree is benchmark hacking facilitated by prompt hacking~\cite{qiu2026benchmark}?
\end{tcolorbox}

To answer these questions, we conduct an empirical study across 11 diverse tasks types including retrieval, classification, clustering, and semantic textual similarity (STS). For each task, we synthesize a set of $~$40 task-specific prompts with different LLMs, removed duplicates, and evaluate a suite of different open-weight embedding models. Our findings reveal that reported benchmark scores frequently misrepresent expected performance: the default prompt can act as an outlier, artificially deflating or inflating a model's perceived quality (\autoref{fig:abstract}). More critically, we show that by adversarially selecting prompts, any model in our study can be made to appear as the top-ranked model on a simulated leaderboard.

We recommend that benchmarks transition from single-point estimates to multi-prompt evaluation — reporting score distributions or robustness metrics alongside point estimates — and that model developers either optimize for prompt robustness or adopt limited prompt schemes such as those used by the Jina v5 series \cite{jina-v5}.

\section{Methodology}

To test the influence of the prompts on the benchmark performance, we selected the following diverse list of tasks and datasets from \texttt{mteb}\footnote{We use the dataset name in \texttt{mteb} to ensure reproducibility.}: retrieval (MIRACLRetrievalHardNegatives.v2, Touche2020Retrieval.v3, FEVERHardNegatives \cite{MIRACL, Thakur_etal_SIGIR2024, thorne-etal-2018-fever}), classification (TweetSentimentClassification, ImdbClassification, AmazonCounterfactualClassification \cite{barbieri-etal-2022-xlm, maas-etal-2011-learning, oneill-etal-2021-wish}), clustering (MedrxivClusteringP2P.v2, StackExchangeClustering.v2 \cite{enevoldsen2025mmtebmassivemultilingualtext, geigle:2021:arxiv}), and semantic similarity (STS15, STS14, STS22.v2 \cite{bicici-2015-rtm, bandhakavi-etal-2014-generating, chen-etal-2022-semeval}). We chose tasks either derived from MTEB \cite{muennighoff2022mteb} or MMTEB \cite{enevoldsen2025mmtebmassivemultilingualtext}. To allow for comparison to the English only BGE models we only evaluate on the English subsets.

For each task, we generated a list of 15 prompts with gpt-oss-120b \cite{openai2025gptoss120bgptoss20bmodel}, Qwen3.6-35B-A3B \cite{qwen36_35b_a3b}, and Gemma 4 26B \cite{gemmateam2026gemma4} with vLLM \cite{kwon2023efficient} structured outputs with the default parameters. Models were instructed to generate prompts based on the combination of language, task description, and task name. The prompts were manually evaluated to ensure that they were coherent and relevant for task (see \autoref{app:prompt_eval}, and \autoref{app:default_prompts} for default prompts). The prompts generation was set up to resemble a real use-case, where the practitioner will rely on a custom, task-oriented prompt. We use synthetic prompt generation to reflect real practitioner behaviour and to avoid author bias in the reported effects. The final list of prompts was deduplicated based on Levenshtein distance and threshlod of 0.1. Number of prompts per task is provided in \autoref{tab:number-of-prompts}. The resulting prompts are short (mean 7.78 words, std 3.2), consistent with documented model usage examples. 
We validated that synthetic and human-written prompts produce largely comparable score distributions, and that where differences exist they do not affect our conclusions (see \autoref{app:prompt_eval}).

\begin{table}[ht]
\centering
\begin{tabular}{lr}
\hline
\textbf{Task} & \textbf{N} \\
\hline
AmazonCounterfactualClassification & 40 \\
FEVERHardNegatives & 39 \\
STS22.v2 & 43 \\
Touche2020Retrieval.v3 & 38 \\
STS15 & 44 \\
STS14 & 45 \\
MIRACLRetrievalHardNegatives.v2 & 39 \\
TweetSentimentClassification & 42 \\
MedrxivClusteringP2P.v2 & 39 \\
StackExchangeClustering.v2 & 40 \\
ImdbClassification & 44 \\
\hline
\end{tabular}
\caption{Number of generated prompts used for each task after deduplication.}
\label{tab:number-of-prompts}
\end{table}

We seek to evaluate a broad selection including both multilingual models: Qwen3-Embedding-0.6B \cite{qwen3embedding}, multilingual-e5-large-instruct \cite{wang2024multilingual}, KaLM-embedding-multilingual-mini-instruct-v2.5 \cite{hu2025kalmembedding}, as well as English models across a variety of sizes; bge-small-en-v1.5, bge-base-en-v1.5, and bge-large-en-v1.5 \cite{bge_embedding}. We ran the models on each task for each of the task-specific prompts, and compared the results with the reported scores on \texttt{mteb}. Models revisions and versions are available in \autoref{app:models_rev}.

\section{Analysis and Discussion}

\begin{figure}
    \centering
    \includegraphics[width=\linewidth]{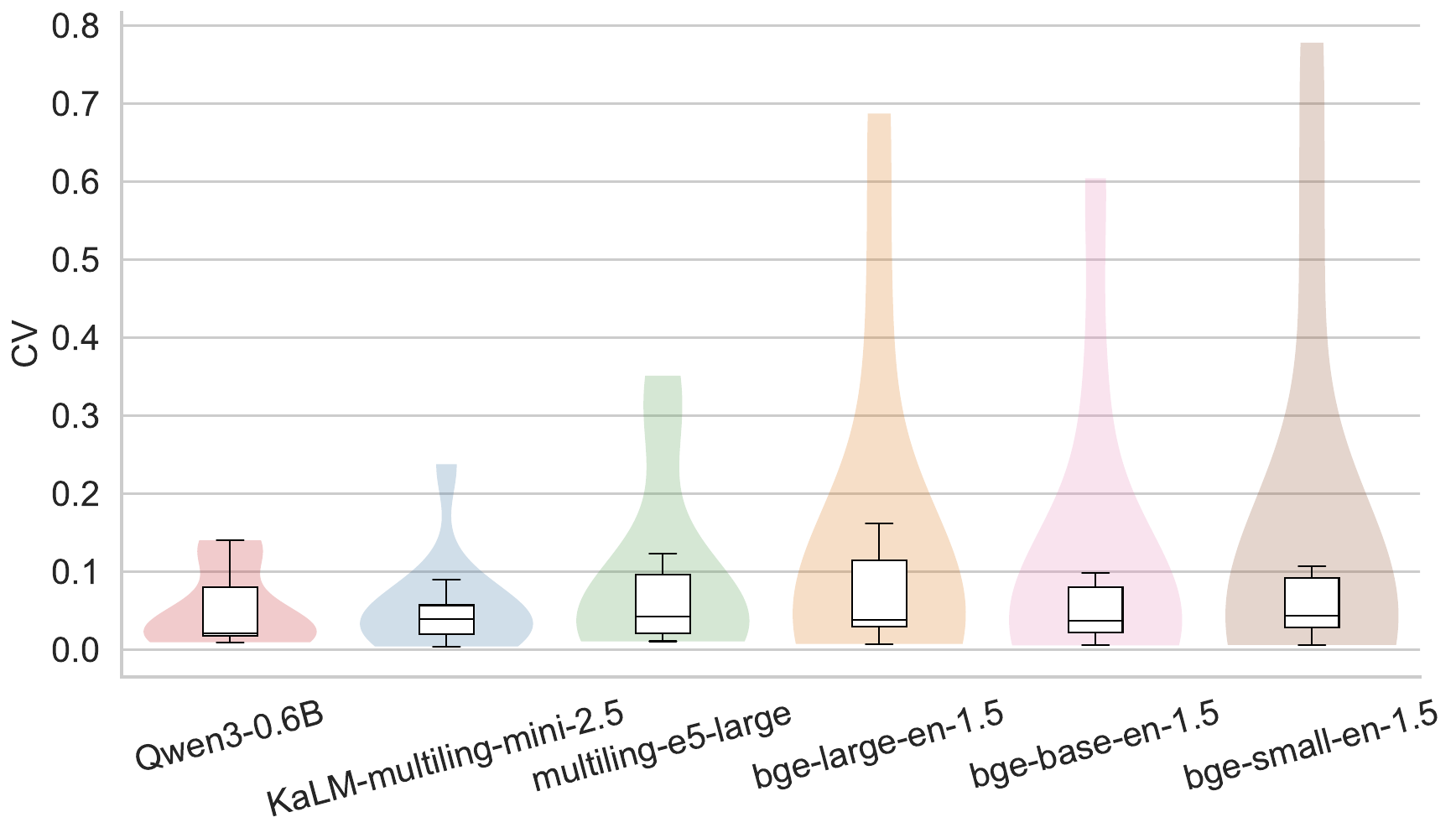}
    \caption{\textbf{Prompt sensitivity}: Coefficient of variance distributions across the models and prompts. We see that certain models have a high degree of variance across prompts.}
    \label{fig:cv_results}
\end{figure}

\subsection{Prompt Inflation and Deflation}
We examine whether the reported single-prompt results are representative of a model's expected performance. In the \autoref{sec:det_scores} (\autoref{fig:app_results}), the score distributions across the prompts for each model/task combination are presented, with the reported \texttt{mteb} score. In \autoref{app:cis_ranks}, the analysis of confidence intervals is presented.

We observe two systematic patterns. First, several models exhibit \textit{prompt deflation}: the reported score is much lower that what would be expected. For example, BGE models show prompt deflation on TweetSentimentClassification, AmazonCounterfactualClassification, and  STS22.v2 tasks (see \autoref{fig:prompt_infl_defl_example}), where its reported scores are in the lower tail of the distributions. In a few cases we even see the reported MTEB score is drastically below the main distribution. This suggests that the default prompts used for these models on these tasks are suboptimal and does not correspond to the expected performance.

Second, we observe \textit{prompt inflation}: the reported score falls in the upper tail of the distribution, overstating expected performance. For example, the reported score of Qwen3 on MIRACL (see \autoref{fig:prompt_infl_defl_example}) task aligns with a secondary mode in the upper range of the distribution, which is above the primary peak. This indicates that the default prompt produces an atypically strong result.

\begin{figure*}[h]
    \centering
    \includegraphics[width=\linewidth]{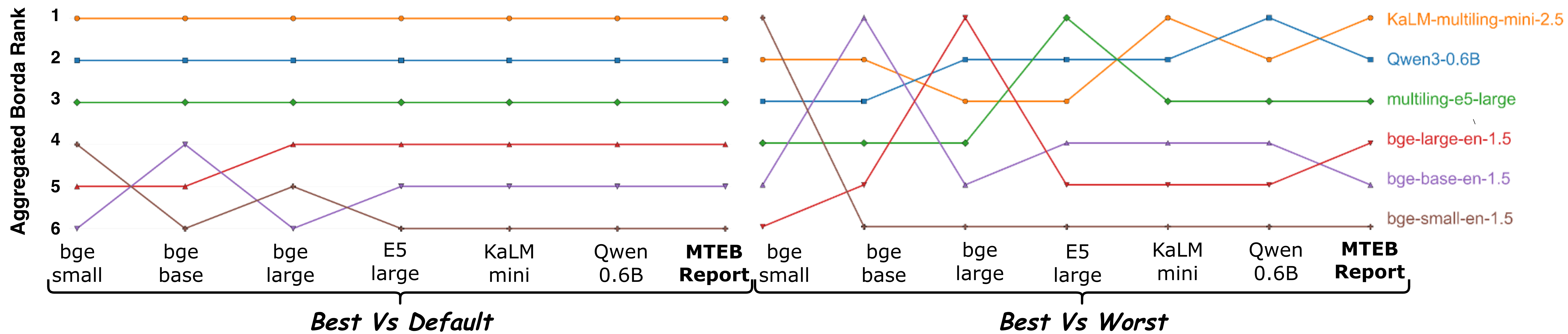}
    \caption{\textbf{Leaderboard sensitivity under adversarial prompt selection}. We show for each model two scenarios. Left: The model uses its best prompt while others use the default. Right: The model uses its best prompt while others use their worst. Ranks are computed via Borda scores across all tasks. In both cases we compare against the default configuration (MTEB Reported).}
    \label{fig:rank_results}
\end{figure*}

\begin{figure*}
    \centering
    \includegraphics[width=\linewidth]{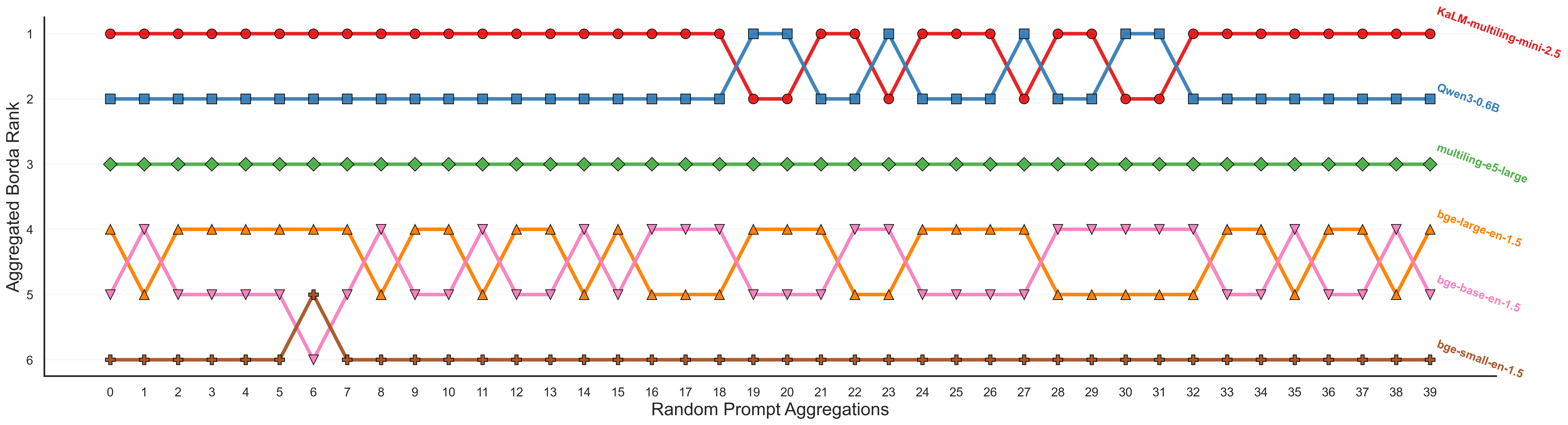}
    \caption{Leaderboard sensitivity under random prompt selection.}
    \label{fig:random_prompts}
\end{figure*}

\subsection{Prompt Sensitivity Analysis}

To answer RQ1, we analyze distributions of scores and coefficients of variation (CVs) of the scores obtained by the tasks per model (see \autoref{fig:cv_results}). The higher the CV is, the less robust the model is for variations in the prompt. We chose CV as it is unit-free, which allows us to compare scores for different tasks. While median CVs are moderate (2--6\%), the distributions exhibit long tails: for the BGE models, individual task CVs reach 30--45\%, confirming that specific model--task combinations are highly sensitive to prompt choice.

\subsection{Effect on Leaderboard Rankings}

Following RQ2, we investigate whether prompt sensitivity can be exploited to alter model rankings. Since \texttt{mteb} allows model contributors to specify custom prompts, a model developer could select a prompt that maximizes their model's score. To simulate this, we construct adversarial leaderboard configurations: for each target model, we assign it its best-performing prompt across tasks while assigning all other models either their worst-performing or default prompts, and compute the aggregated Borda rank. The rank changes per configuration are presented on \autoref{fig:rank_results}.

Under default prompts, KaLM and Qwen3 occupy ranks 1–2, e5-large rank 3, and the BGE variants ranks 4–6. A well-meaning developer who simply reports the prompt that worked best during development can therefore distort the leaderboard. However, under adversarial selection (an upper bound on manipulability), every model in our study can be promoted to rank 1, including bge-small, which rises from last place. This configuration does not model public submissions, where each contributor sets only their own prompt; rather, it corresponds to a setting where one party controls all prompt assignments, such as a selective write-up or public release, and it bounds how far favorable prompt selection can move a ranking.

Importantly, the milder scenario with the best performing prompt for the target model and default for others, still produces meaningful rank changes. For example, bge-large rises from rank 5 to rank 3 under this configuration, overtaking e5-large. This demonstrates that even a single model contributor optimizing their prompt, without any adversarial intent toward competitors, can distort the leaderboard.

We run the leaderboard sensitivity analysis sampling random prompts as well. The results are present in \autoref{fig:random_prompts}. They support our prior findings and display similar trends when it comes to the rankings.

\subsection{Prompt Inflation in Reported Scores}
\begin{figure}[h]
    \centering
    \includegraphics[width=\linewidth]{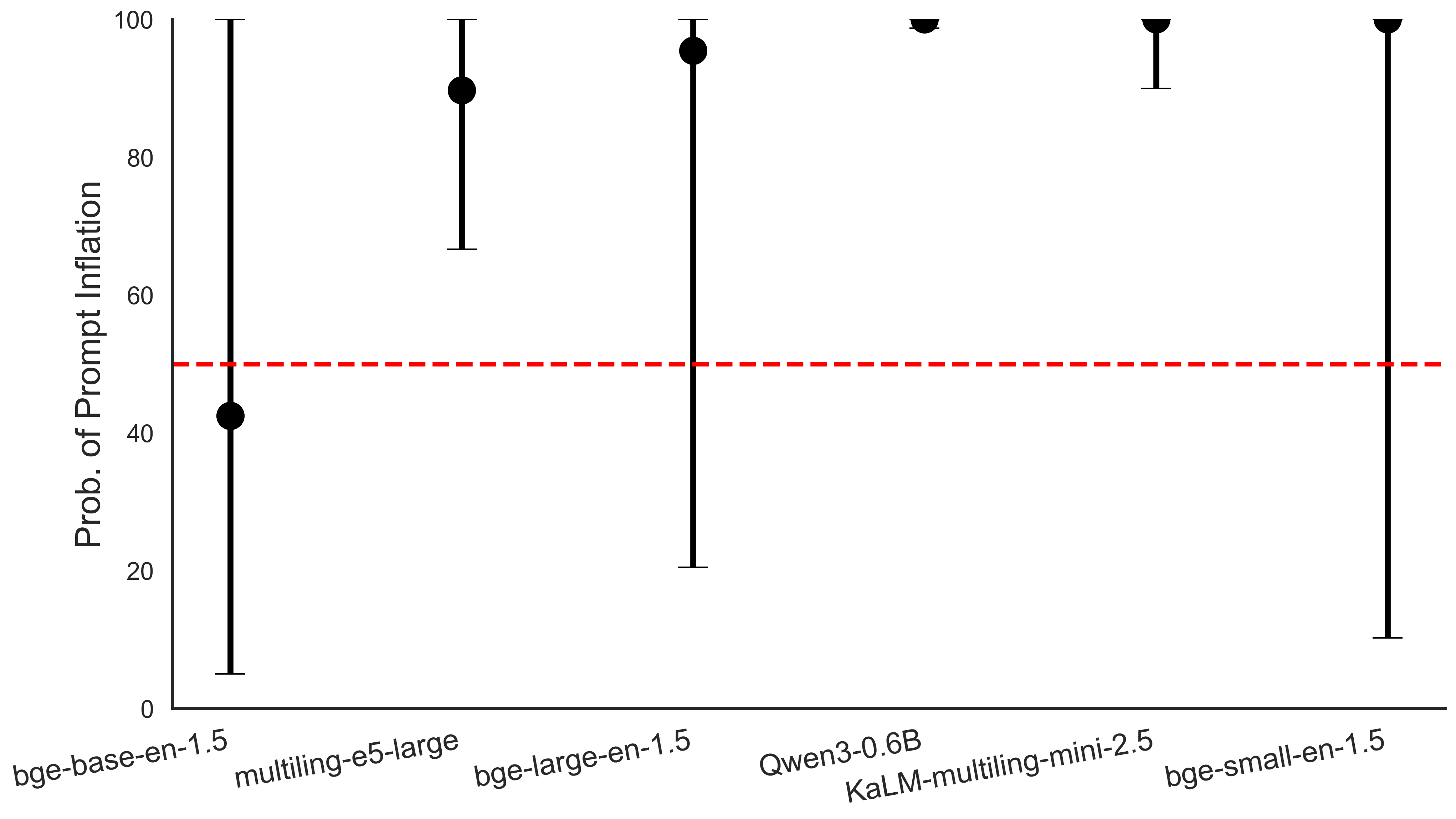}
    \caption{\textbf{Probability of prompt inflation}, i.e. reported score higher than the median, across tasks. The uncertainty is a 95\% bootstrapped confidence interval.}
    \label{fig:boot_medians}
\end{figure}

Lastly, to examine whether some models systematically overfit (RQ3), we analyzed how the reported MTEB scores correspond to an observed score distribution. For each model--task combination, we calculated the empirical probability of the reported score across obtained scores (see \autoref{fig:boot_medians}). Our results reveal a systematic issue: most models report scores above the median of the prompt distribution, potentially of the intended instruction. It can be interpreted as a systematic gap between the single reported estimate and the performance expected across plausible prompts. This pattern is analogous to p-hacking \cite{simmons2011} in that it exploits undisclosed degrees of freedom in evaluation without constituting an explicit violation of submission policies — we term this \textit{prompt hacking} \cite{qiu2026benchmark}: the selective reporting of a single favorable prompt while remaining compliant, without modifying model weights, training data, or architecture.
Importantly, prompt hacking need not be intentional: just as analytical bias can arise from well-meaning researchers exploring analytical choices without recognizing the inflationary effect \cite{gelman2016statistical}, a model developer may simply report the prompt that worked best during development without deliberate intent to game the benchmark.

\section{Conclusion}

In this paper, we conducted an empirical study of prompt sensitivity across 6 models and 11 tasks. We find that reported benchmark scores frequently misrepresent expected performance: the MTEB default prompt can systematically overstate or understate a model's quality. Under adversarial prompt selection, every model in our study can be promoted to rank 1, including models that rank last under default evaluation. Even a single developer optimizing their prompt can shift rankings by multiple positions. We recommend that benchmarks adopt multi-prompt evaluation or similar strategies to ensure that the reported rankings faithfully reflect the models' quality.

\section*{Limitations}
\textbf{Model scope:} We evaluate 6 open-weight models with up to 0.6B parameters. Larger instruction-tuned embedding models and closed-source models (e.g., OpenAI text-embedding-3, Cohere embed-v4) may exhibit different sensitivity profiles.

\noindent\textbf{Prompt generation and sample size:} Our prompt sets consist of $\sim$40 synthetically generated prompts per task, produced by a single language model. This is a relatively small sample: variance estimates (e.g., CV) computed over $\sim$40 points carry non-trivial uncertainty, and the synthetic generation process may not fully capture the diversity of prompts that practitioners write in practice. A larger, more diverse prompt set would yield more robust conclusions.

\noindent\textbf{Language coverage:} We restrict our analysis to English. Multilingual settings may introduce additional sources of prompt sensitivity related to language-specific phrasing conventions and translation artifacts.

\noindent\textbf{Prompt variability as a practitioner asset:}
Our study frames prompt sensitivity primarily as a threat to benchmark validity, but the same variability can be an asset in practice. A developer who tries even a small set of task-specific prompts — as our results show — can gain up to 10 percentage points in performance at essentially no cost. This work does not evaluate strategies for exploiting this, such as prompt search or ensembling, and the optimal number of prompts to try in a practical setting remains an open question.

\newpage

\section*{Acknowledgments}

Yevhen Kostiuk and Kenneth Enevoldsen are funded by the Danish Foundation Models project (4378-00001B). Kenneth Enevoldsen is additionally funded by the European Union, Horizon Europe (101178170), the Danish National Research Foundation (DNRF193), the Aage and Johanne Louis-Hansens Foundation (25-1-17733), and the Augustinus Foundation (2025-0299).

Part of the computation done for this project was performed on the UCloud interactive HPC system, which is managed by the eScience Center at the University of Southern Denmark.

\bibliography{custom}

\newpage
\appendix

\section{Detailed Scores} \label{sec:det_scores}

\begin{figure*}
    \centering

    \includegraphics[width=\linewidth]{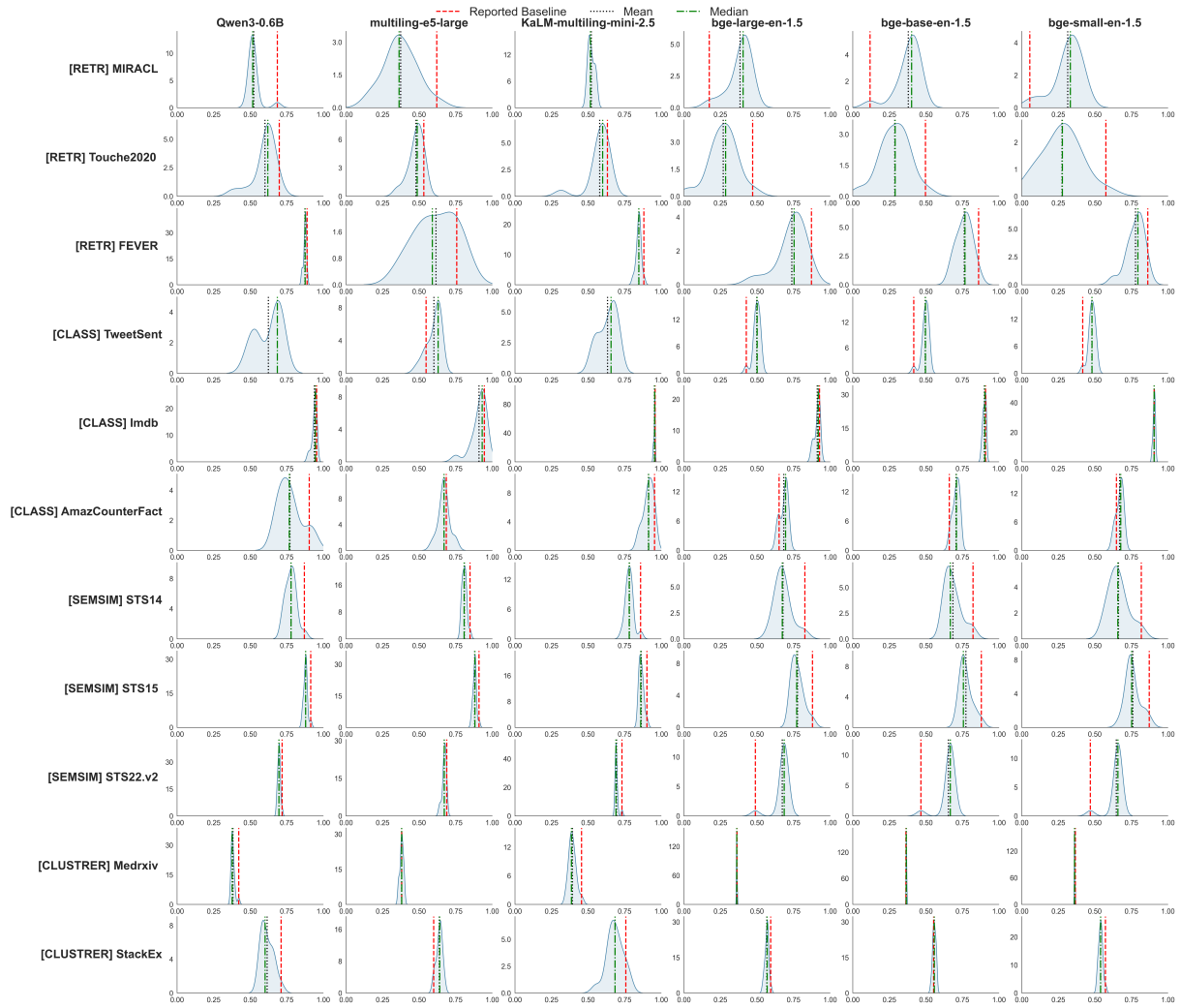}
    \caption{Detailed scores per model, metric, and task. For some tasks MTEB does not currently report a score hence the red line is missing.}
    \label{fig:app_results}
\end{figure*}

\section{Model Revisions} \label{app:models_rev}
For our experiments, we used the following model and revisions from HuggingFace Hub (see \autoref{tab:model_revisions}) 
\cite{wolf2020huggingfacestransformersstateoftheartnatural} and MTEB version 2.12.30.

\begin{table*}
\centering
\small
\begin{tabular}{ll}
\toprule
\textbf{Model} & \textbf{Commit} \\
\midrule
\href{https://huggingface.co/Qwen/Qwen3-Embedding-0.6B}{Qwen/Qwen3-Embedding-0.6B} 
    & \texttt{97b0c614} \\
\href{https://huggingface.co/intfloat/multilingual-e5-large-instruct}{intfloat/multilingual-e5-large-instruct} 
    & \texttt{274baa43} \\
\href{https://huggingface.co/KaLM-Embedding/KaLM-embedding-multilingual-mini-instruct-v2.5}{KaLM-Embedding/KaLM-embedding-multilingual-mini-instruct-v2.5} 
    & \texttt{52c687bb} \\
\href{https://huggingface.co/BAAI/bge-small-en-v1.5}{BAAI/bge-small-en-v1.5} 
    & \texttt{5c38ec7c} \\
\href{https://huggingface.co/BAAI/bge-base-en-v1.5}{BAAI/bge-base-en-v1.5} 
    & \texttt{a5beb1e3} \\
\href{https://huggingface.co/BAAI/bge-large-en-v1.5}{BAAI/bge-large-en-v1.5} 
    & \texttt{d4aa6901} \\
\bottomrule
\end{tabular}
\caption{Model revisions used in experiments.}
\label{tab:model_revisions}
\end{table*}

\section{Confidence Intervals for Borda Ranks} \label{app:cis_ranks}

\begin{figure*}
    \centering
    \includegraphics[width=\linewidth]{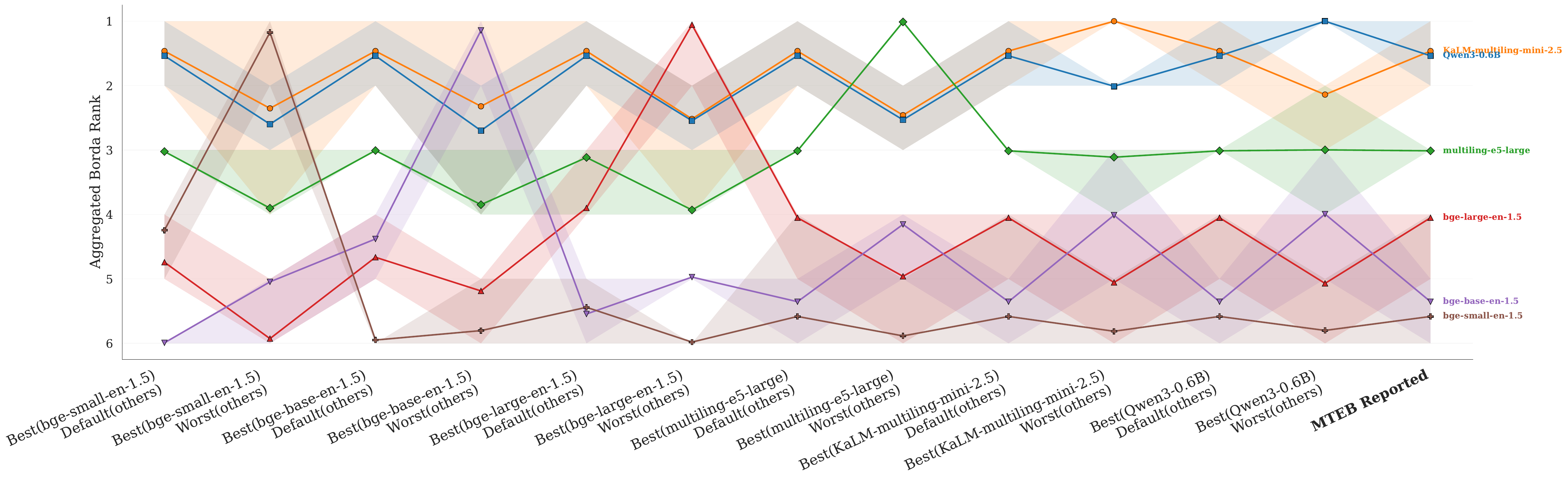}
    \caption{CIs of the ranks.}
    \label{fig:cis_borda_ranks}
\end{figure*}

\begin{table}[t]
\centering
\small
\setlength{\tabcolsep}{4pt}
\begin{tabular}{l l r r}
\toprule
\textbf{Model} & \textbf{Cond.} & \textbf{Rank} & \textbf{Borda} \\
\midrule
Qwen3-0.6B & S-D & 1.537 [1--2] & 4.397 [4.13--4.64] \\
KaLM & S-D & 1.463 [1--2] & 4.451 [4.10--4.80] \\
multiling-e5-large & S-D & 3.023 [3--3] & 2.636 [2.09--3.09] \\
bge-large-en-1.5 & S-D & 4.741 [4--5] & 1.281 [0.82--1.73] \\
bge-base-en-1.5 & S-D & 5.992 [6--6] & 0.540 [0.27--0.82] \\
bge-small-en-1.5 & S-D & 4.244 [4--5] & 1.633 [1.18--2.09] \\
\midrule
Qwen3-0.6B & S-W & 2.598 [2--3] & 3.283 [2.82--3.73] \\
KaLM & S-W & 2.353 [1--4] & 3.371 [2.56--4.20] \\
multiling-e5-large & S-W & 3.902 [3--4] & 2.273 [1.55--3.00] \\
bge-large-en-1.5 & S-W & 5.927 [5--6] & 0.554 [0.27--0.82] \\
bge-base-en-1.5 & S-W & 5.046 [5--6] & 1.105 [0.64--1.55] \\
bge-small-en-1.5 & S-W & 1.174 [1--2] & 4.065 [3.64--4.45] \\
\midrule
Qwen3-0.6B & B-D & 1.537 [1--2] & 4.397 [4.13--4.64] \\
KaLM & B-D & 1.463 [1--2] & 4.451 [4.10--4.80] \\
multiling-e5-large & B-D & 3.007 [3--3] & 2.636 [2.09--3.09] \\
bge-large-en-1.5 & B-D & 4.664 [4--5] & 1.371 [1.00--1.82] \\
bge-base-en-1.5 & B-D & 4.380 [4--5] & 1.538 [1.27--1.82] \\
bge-small-en-1.5 & B-D & 5.949 [6--6] & 0.545 [0.09--1.09] \\
\midrule
Qwen3-0.6B & B-W & 2.698 [2--4] & 3.192 [2.78--3.60] \\
KaLM & B-W & 2.321 [1--4] & 3.371 [2.56--4.20] \\
multiling-e5-large & B-W & 3.848 [3--4] & 2.346 [1.64--3.00] \\
bge-large-en-1.5 & B-W & 5.187 [5--6] & 0.919 [0.73--1.09] \\
bge-base-en-1.5 & B-W & 1.143 [1--2] & 4.149 [3.73--4.55] \\
bge-small-en-1.5 & B-W & 5.803 [5--6] & 0.666 [0.18--1.18] \\
\bottomrule
\end{tabular}
\caption{Model rankings and Borda scores for BGE-small and BGE-base prompt conditions.}
\label{tab:results-part1}
\end{table}

\begin{table}[t]
\centering
\small
\setlength{\tabcolsep}{4pt}
\begin{tabular}{l l r r}
\toprule
\textbf{Model} & \textbf{Cond.} & \textbf{Rank} & \textbf{Borda} \\
\midrule
Qwen3-0.6B & L-D & 1.537 [1--2] & 4.397 [4.13--4.64] \\
KaLM & L-D & 1.463 [1--2] & 4.451 [4.10--4.80] \\
multiling-e5-large & L-D & 3.115 [3--4] & 2.540 [2.00--3.00] \\
bge-large-en-1.5 & L-D & 3.899 [3--4] & 2.006 [1.55--2.45] \\
bge-base-en-1.5 & L-D & 5.545 [5--6] & 0.724 [0.45--0.91] \\
bge-small-en-1.5 & L-D & 5.441 [5--6] & 0.820 [0.36--1.36] \\
\midrule
Qwen3-0.6B & L-W & 2.545 [2--3] & 3.237 [2.80--3.70] \\
KaLM & L-W & 2.517 [2--4] & 3.232 [2.44--4.00] \\
multiling-e5-large & L-W & 3.930 [3--4] & 2.145 [1.45--2.91] \\
bge-large-en-1.5 & L-W & 1.055 [1--2] & 4.232 [3.82--4.64] \\
bge-base-en-1.5 & L-W & 4.972 [5--5] & 1.197 [0.82--1.55] \\
bge-small-en-1.5 & L-W & 5.981 [6--6] & 0.581 [0.27--0.91] \\
\midrule
Qwen3-0.6B & E-D & 1.537 [1--2] & 4.397 [4.13--4.64] \\
KaLM & E-D & 1.463 [1--2] & 4.451 [4.10--4.80] \\
multiling-e5-large & E-D & 3.012 [3--3] & 2.726 [2.18--3.18] \\
bge-large-en-1.5 & E-D & 4.051 [4--5] & 1.643 [1.18--2.09] \\
bge-base-en-1.5 & E-D & 5.354 [5--6] & 0.901 [0.64--1.09] \\
bge-small-en-1.5 & E-D & 5.583 [4--6] & 0.820 [0.36--1.36] \\
\midrule
Qwen3-0.6B & E-W & 2.531 [2--3] & 3.450 [3.11--3.80] \\
KaLM & E-W & 2.459 [2--3] & 3.481 [2.78--4.00] \\
multiling-e5-large & E-W & 1.011 [1--1] & 4.544 [4.18--4.82] \\
bge-large-en-1.5 & E-W & 4.960 [4--6] & 1.064 [0.73--1.45] \\
bge-base-en-1.5 & E-W & 4.154 [4--5] & 1.527 [1.09--1.91] \\
bge-small-en-1.5 & E-W & 5.885 [5--6] & 0.619 [0.27--1.00] \\
\midrule
Qwen3-0.6B & K-D & 1.537 [1--2] & 4.397 [4.13--4.64] \\
KaLM & K-D & 1.463 [1--2] & 4.451 [4.10--4.80] \\
multiling-e5-large & K-D & 3.012 [3--3] & 2.726 [2.18--3.18] \\
bge-large-en-1.5 & K-D & 4.051 [4--5] & 1.643 [1.18--2.09] \\
bge-base-en-1.5 & K-D & 5.354 [5--6] & 0.901 [0.64--1.09] \\
bge-small-en-1.5 & K-D & 5.583 [4--6] & 0.820 [0.36--1.36] \\
\midrule
Qwen3-0.6B & K-W & 2.010 [2--2] & 3.693 [3.44--3.91] \\
KaLM & K-W & 1.000 [1--1] & 5.000 [5.00--5.00] \\
multiling-e5-large & K-W & 3.111 [3--4] & 2.479 [1.82--3.09] \\
bge-large-en-1.5 & K-W & 5.055 [4--6] & 1.182 [0.82--1.55] \\
bge-base-en-1.5 & K-W & 4.011 [3--5] & 1.800 [1.36--2.19] \\
bge-small-en-1.5 & K-W & 5.813 [5--6] & 0.818 [0.36--1.36] \\
\midrule
Qwen3-0.6B & Q-D & 1.537 [1--2] & 4.397 [4.13--4.64] \\
KaLM & Q-D & 1.463 [1--2] & 4.451 [4.10--4.80] \\
multiling-e5-large & Q-D & 3.012 [3--3] & 2.726 [2.18--3.18] \\
bge-large-en-1.5 & Q-D & 4.051 [4--5] & 1.643 [1.18--2.09] \\
bge-base-en-1.5 & Q-D & 5.354 [5--6] & 0.901 [0.64--1.09] \\
bge-small-en-1.5 & Q-D & 5.583 [4--6] & 0.820 [0.36--1.36] \\
\midrule
Qwen3-0.6B & Q-W & 1.000 [1--1] & 4.902 [4.73--5.00] \\
KaLM & Q-W & 2.139 [2--3] & 3.269 [2.64--3.75] \\
multiling-e5-large & Q-W & 2.997 [2--4] & 2.592 [1.91--3.18] \\
bge-large-en-1.5 & Q-W & 5.070 [4--6] & 1.240 [0.91--1.55] \\
bge-base-en-1.5 & Q-W & 3.994 [3--5] & 1.876 [1.36--2.36] \\
bge-small-en-1.5 & Q-W & 5.800 [5--6] & 0.895 [0.36--1.45] \\
\midrule
Qwen3-0.6B & MTEB & 1.537 [1--2] & 4.397 [4.13--4.64] \\
KaLM & MTEB & 1.463 [1--2] & 4.451 [4.10--4.80] \\
multiling-e5-large & MTEB & 3.012 [3--3] & 2.726 [2.18--3.18] \\
bge-large-en-1.5 & MTEB & 4.051 [4--5] & 1.643 [1.18--2.09] \\
bge-base-en-1.5 & MTEB & 5.354 [5--6] & 0.901 [0.64--1.09] \\
bge-small-en-1.5 & MTEB & 5.583 [4--6] & 0.820 [0.36--1.36] \\
\bottomrule
\end{tabular}
\caption{Model rankings and Borda scores for the remaining prompt conditions.}
\label{tab:results-part2}
\end{table}

To assess the sensitivity of the MTEB leaderboard to prompt selection, we recompute the Borda-count aggregate ranking and quantify the resulting rank uncertainty via a nonparametric bootstrap. To estimate confidence intervals (CIs) on the resulting ranks, we bootstrap over two variations: (1) the set of evaluation tasks, resampled with replacement to reflect uncertainty in benchmark composition, and (2) the set of sampled prompts per task-model pair, resampled with replacement (excluding the fixed default prompt) to reflect the fact that the observed best/worst scores are themselves noisy extreme-value estimates drawn from a finite prompt sample. We repeat this joint resampling procedure 1,000 times, recomputing the full Borda ranking on each replicate, and report the mean rank together with the 5th and 95th percentiles of the bootstrap rank distribution as a 90\% confidence interval for each model under each regime.

The results are presented in \autoref{fig:cis_borda_ranks}, \autoref{tab:results-part}, and \autoref{tab:results-part2}. Under the reported (reportedprompt) leaderboard, models separate into three loose tiers — KaLM-multiling-mini-2.5 and Qwen3-0.6B occupy the top two positions (mean ranks 1.46 and 1.54, 90\% CI [1,2] for both), multiling-e5-large sits alone in third (mean rank 3.01, CI [3,3]), and bge-large-en-1.5, bge-base-en-1.5, and bge-small-en-1.5 trail behind, with bge-base and bge-small's intervals overlapping ([5,6] and [4,6] respectively). The overlapping intervals at both the top and bottom of the table indicate that the reported ordering within these pairs is not statistically well-separated. This finding futher supports the claim that ordering is highly influenced by prompt selection: when a model is granted its best prompt while competitors are forced to use their worst, every one of the six models is able to reach or approach rank 1. Notably, KaLM-multiling-mini-2.5 and Qwen3-0.6B are the only two models that reach rank 1 with zero-width confidence intervals ([1,1]) under their respective best-self/worst-others regimes, suggesting their advantage over the remaining four models is comparatively robust to prompt choice even though the top-two ordering between themselves is not. In contrast, the best prompt for one model against competitors' reported scores produces far smaller rank shifts for larger models (KaLM and Qwen3's), showing that most of the observed leaderboard instability stems from the choice of prompt used for competing models rather than from the model's own prompt optimization alone.

\section{Prompt Evaluation} \label{app:prompt_eval}
We manually evaluated 100 generated prompts. For each prompt, we looked at the task and task description, and assigned one of the following labels:
\begin{itemize}
    \item Acceptable. The prompt sounds natural, relevant to the task, and the task is clearly stated.
    \item Irrelevant. The prompt is irrelevant to the task.
    \item Unnatural. The prompt does not sound correct and is not natural.
\end{itemize}

Based on our analysis \textit{all} the generated prompts were considered to be acceptable. We noticed that some prompts contained name of the task in them, e.g. ``Similarity score request, eng, SemEval STS 2014:''. We marked these prompts as acceptable, as such prompts contain reasonable instructions, and the task name may provide an indication of the task to the model. In the case of the example, SemEval is a well known shared task, therefore the model may potentially know its relation towards STS.

\begin{figure*}
    \centering
    \includegraphics[width=0.7\linewidth]{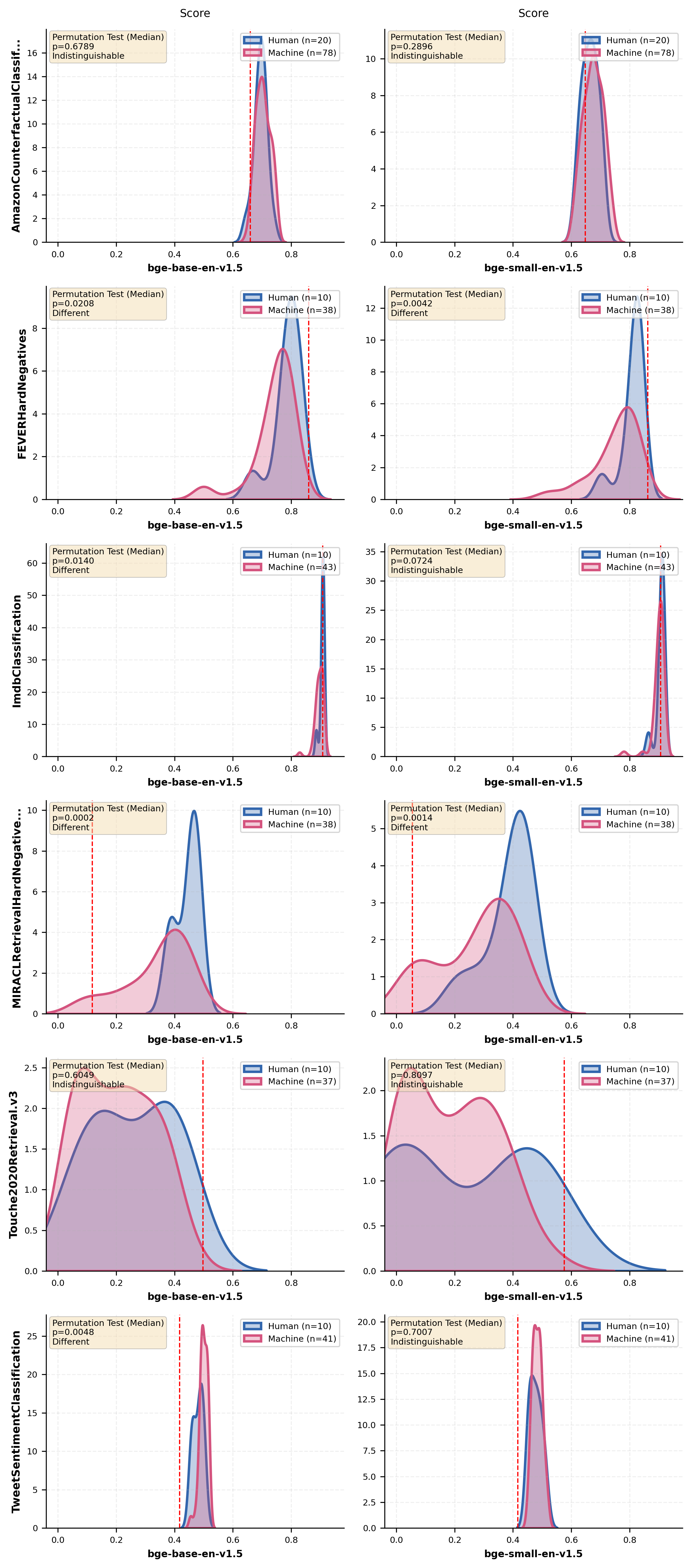}
    \caption{Comparison of human vs machine generated prompt score distribution for Classification and Retrieval tasks. Red dashed line indicates a reported score on MTEB.}
    \label{fig:app_human_vs_machine_1}
\end{figure*}

\begin{figure*}
    \centering
    \includegraphics[width=0.7\linewidth]{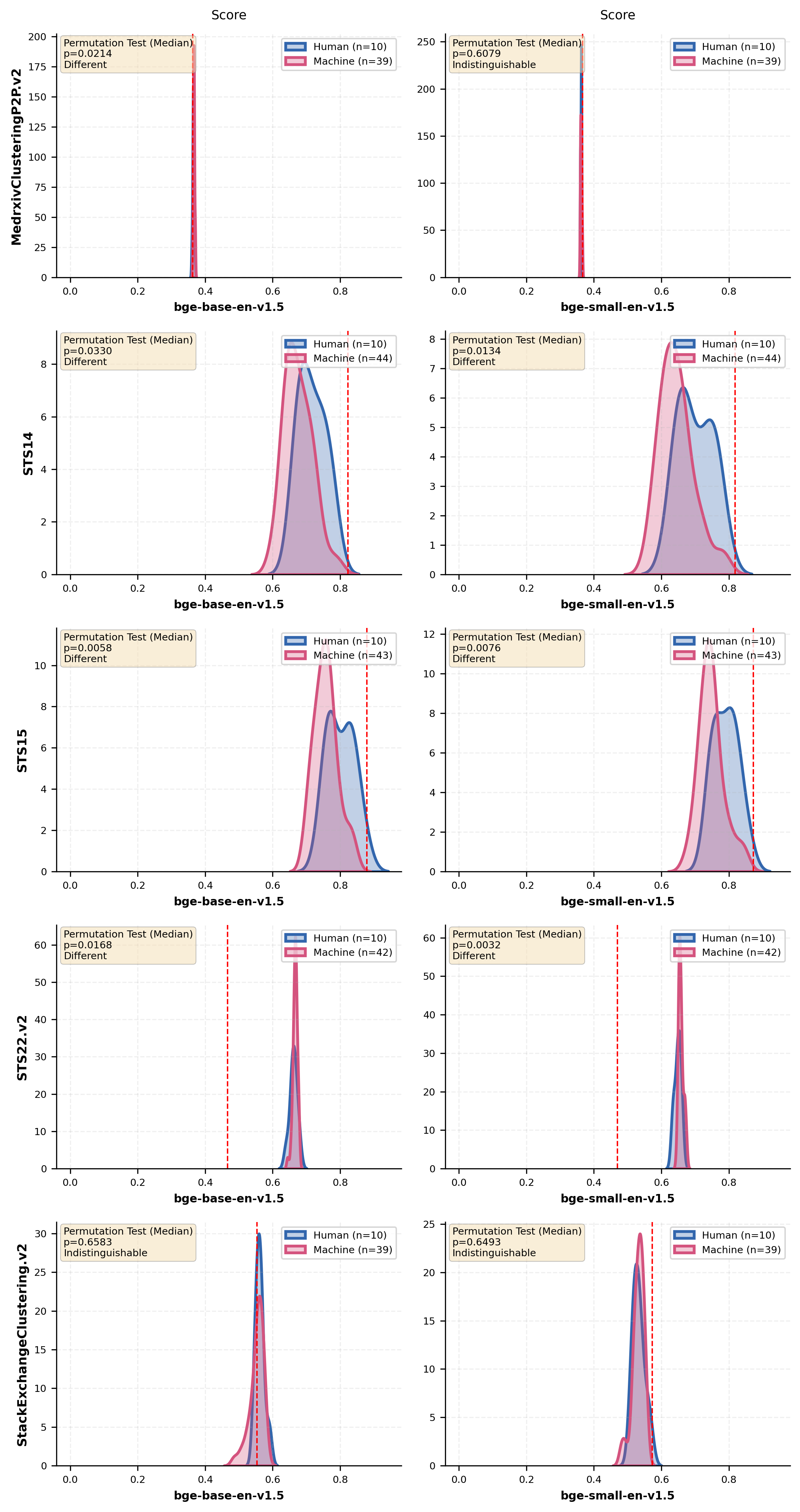}
    \caption{Comparison of human vs machine generated prompt score distribution for Clustering and STS tasks. Red dashed line indicates a reported score on MTEB.}
    \label{fig:app_human_vs_machine_2}
\end{figure*}

To show that our findings are generalizable to natural human prompts, we created 10 prompts per task for BGE family of models and compared the human prompts performance with the machine generated prompts. The results are presented on \autoref{fig:app_human_vs_machine_1} and \autoref{fig:app_human_vs_machine_2}. The distributions were tested with the permutation test on similarity of distributions.

The figure shows that the human and synthetic prompts perform similarly in most cases. In the cases, where the test did not reject the hypothesis of distributional similarity, the analysis of the reported score and performance distribution does not contradict findings of the paper. 

The acceptability check was carried out by a single annotator (one of the authors). A limitation of this approach is that a single annotator may hold an implicit bias toward accepting the generated prompts. The scheme used is a coarse coherence filter (Acceptable / Irrelevant / Unnatural) that targets clearly broken instructions rather than fine-grained quality, which limits the room for subjective disagreement.  The annotation results excluded nothing: all prompts were judged Acceptable, so no regeneration or replacement was performed and no borderline decision affected the prompt set. Had any prompt been rejected, our policy was to remove rejected prompts for that task to regenerate the same number of missing prompts and evaluate them again, until a full set of acceptable prompts was obtained.

\section{Default MTEB Prompts} \label{app:default_prompts}

In \autoref{tab:default_model_task_prompts}, \autoref{tab:default_model_task_prompts_1}, and \autoref{tab:default_model_task_prompts_2} the default MTEB prompts are present per each model--task combination.

\begin{table*}
\centering
\small
\caption{Default prompts by model and task.}
\begin{tabular}{|p{3cm}|p{3cm}|p{7cm}|}
\hline
\textbf{Model} & \textbf{Task} & \textbf{Prompt} \\
\hline
Qwen/Qwen3-Embedding-0.6B & AmazonCounterfactual & Classify a given Amazon customer review text as either counterfactual or not-counterfactual \\
\hline
Qwen/Qwen3-Embedding-0.6B & FEVERHardNegatives & Retrieve text based on user query. \\
\hline
Qwen/Qwen3-Embedding-0.6B & STS22.v2 & Retrieve semantically similar text. \\
\hline
Qwen/Qwen3-Embedding-0.6B & Touche2020Retrieval.v3 & Retrieve text based on user query. \\
\hline
Qwen/Qwen3-Embedding-0.6B & STS15 & Retrieve semantically similar text. \\
\hline
Qwen/Qwen3-Embedding-0.6B & STS14 & Retrieve semantically similar text. \\
\hline
Qwen/Qwen3-Embedding-0.6B & MIRACLRetrievalHard & Given a question, retrieve Wikipedia passages that answer the question \\
\hline
Qwen/Qwen3-Embedding-0.6B & TweetSentiment & Classify user passages. \\
\hline
Qwen/Qwen3-Embedding-0.6B & MedrxivClusteringP2P.v2 & Identify the main category of Medrxiv papers based on the titles and abstracts \\
\hline
Qwen/Qwen3-Embedding-0.6B & StackExchangeClustering.v & Identify the topic or theme of StackExchange posts based on the titles \\
\hline
Qwen/Qwen3-Embedding-0.6B & ImdbClassification & Classify the sentiment expressed in the given movie review text from the IMDB dataset \\
\hline
BAAI/bge-small-en-v1.5 & AmazonCounterfactual & Represent this sentence for searching relevant passages:  \\
\hline
BAAI/bge-small-en-v1.5 & FEVERHardNegatives & Represent this sentence for searching relevant passages:  \\
\hline
BAAI/bge-small-en-v1.5 & STS22.v2 & Represent this sentence for searching relevant passages:  \\
\hline
BAAI/bge-small-en-v1.5 & Touche2020Retrieval.v3 & Represent this sentence for searching relevant passages:  \\
\hline
BAAI/bge-small-en-v1.5 & STS15 & Represent this sentence for searching relevant passages:  \\
\hline
BAAI/bge-small-en-v1.5 & STS14 & Represent this sentence for searching relevant passages:  \\
\hline
BAAI/bge-small-en-v1.5 & MIRACLRetrievalHard & Represent this sentence for searching relevant passages:  \\
\hline
BAAI/bge-small-en-v1.5 & TweetSentiment & Represent this sentence for searching relevant passages:  \\
\hline
BAAI/bge-small-en-v1.5 & MedrxivClusteringP2P.v2 & Represent this sentence for searching relevant passages:  \\
\hline
BAAI/bge-small-en-v1.5 & StackExchangeClustering & Represent this sentence for searching relevant passages:  \\
\hline
BAAI/bge-small-en-v1.5 & ImdbClassification & Represent this sentence for searching relevant passages:  \\
\hline
BAAI/bge-base-en-v1.5 & AmazonCounterfactual & Represent this sentence for searching relevant passages:  \\
\hline
BAAI/bge-base-en-v1.5 & FEVERHardNegatives & Represent this sentence for searching relevant passages:  \\
\hline
BAAI/bge-base-en-v1.5 & STS22.v2 & Represent this sentence for searching relevant passages: \\
\hline
\end{tabular}
\label{tab:default_model_task_prompts}
\end{table*}

\begin{table*}
\centering
\small
\caption{Default prompts by model and task.}
\begin{tabular}{|p{3cm}|p{3cm}|p{7cm}|}
\hline
\textbf{Model} & \textbf{Task} & \textbf{Prompt} \\
\hline
BAAI/bge-base-en-v1.5 & Touche2020Retrieval.v3 & Represent this sentence for searching relevant passages:  \\
\hline
BAAI/bge-base-en-v1.5 & STS15 & Represent this sentence for searching relevant passages:  \\
\hline
BAAI/bge-base-en-v1.5 & STS14 & Represent this sentence for searching relevant passages:  \\
\hline
BAAI/bge-base-en-v1.5 & MIRACLRetrievalHard & Represent this sentence for searching relevant passages:  \\
\hline
BAAI/bge-base-en-v1.5 & TweetSentiment & Represent this sentence for searching relevant passages:  \\
\hline
BAAI/bge-base-en-v1.5 & MedrxivClusteringP2P.v2 & Represent this sentence for searching relevant passages:  \\
\hline
BAAI/bge-base-en-v1.5 & StackExchangeClustering.v & Represent this sentence for searching relevant passages:  \\
\hline
BAAI/bge-base-en-v1.5 & ImdbClassification & Represent this sentence for searching relevant passages:  \\
\hline
BAAI/bge-large-en-v1.5 & AmazonCounterfactual & Represent this sentence for searching relevant passages:  \\
\hline
BAAI/bge-large-en-v1.5 & FEVERHardNegatives & Represent this sentence for searching relevant passages:  \\
\hline
BAAI/bge-large-en-v1.5 & STS22.v2 & Represent this sentence for searching relevant passages:  \\
\hline
BAAI/bge-large-en-v1.5 & Touche2020Retrieval.v3 & Represent this sentence for searching relevant passages:  \\
\hline
BAAI/bge-large-en-v1.5 & STS15 & Represent this sentence for searching relevant passages:  \\
\hline
BAAI/bge-large-en-v1.5 & STS14 & Represent this sentence for searching relevant passages:  \\
\hline
BAAI/bge-large-en-v1.5 & MIRACLRetrievalHard & Represent this sentence for searching relevant passages:  \\
\hline
BAAI/bge-large-en-v1.5 & TweetSentiment & Represent this sentence for searching relevant passages:  \\
\hline
BAAI/bge-large-en-v1.5 & MedrxivClusteringP2P.v2 & Represent this sentence for searching relevant passages:  \\
\hline
BAAI/bge-large-en-v1.5 & StackExchangeClustering.v & Represent this sentence for searching relevant passages:  \\
\hline
BAAI/bge-large-en-v1.5 & ImdbClassification & Represent this sentence for searching relevant passages:  \\
\hline
intfloat/multilingual-e5- & AmazonCounterf & Classify a given Amazon customer review text as either counterfactual or not-counterfactual \\
\hline
intfloat/multilingual-e5- & FEVERHardNegatives & Retrieve text based on user query. \\
\hline
intfloat/multilingual-e5- & STS22.v2 & Retrieve semantically similar text. \\
\hline
intfloat/multilingual-e5- & Touche2020Retrieval.v3 & Retrieve text based on user query. \\
\hline
intfloat/multilingual-e5- & STS15 & Retrieve semantically similar text. \\
\hline
intfloat/multilingual-e5- & STS14 & Retrieve semantically similar text. \\
\hline
intfloat/multilingual-e5- & MIRACLRetrievalHard & Given a question, retrieve Wikipedia passages that answer the question \\
\hline
intfloat/multilingual-e5- & TweetSentiment & Classify user passages. \\
\hline
intfloat/multilingual-e5- & MedrxivClusteringP2P.v2 & Identify the main category of Medrxiv papers based on the titles and abstracts \\
\hline
intfloat/multilingual-e5- & StackExchangeClustering & Identify the topic or theme of StackExchange posts based on the titles \\
\hline
intfloat/multilingual-e5- & ImdbClassification & Classify the sentiment expressed in the given movie review text from the IMDB dataset \\
\hline
KaLM-Embedding/KaLM-embed & AmazonCounterfactual & Classify a given Amazon customer review text as either counterfactual or not-counterfactual \\
\hline
\end{tabular}
\label{tab:default_model_task_prompts_1}
\end{table*}

\begin{table*}
\centering
\small
\caption{Default prompts by model and task.}
\begin{tabular}{|p{3cm}|p{3cm}|p{7cm}|}
\hline
\textbf{Model} & \textbf{Task} & \textbf{Prompt} \\
\hline
KaLM-Embedding/KaLM-embed & FEVERHardNegatives & Retrieve text based on user query. \\
\hline
KaLM-Embedding/KaLM-embed & STS22.v2 & Retrieve semantically similar text. \\
\hline
KaLM-Embedding/KaLM-embed & Touche2020Retrieval.v3 & Retrieve text based on user query. \\
\hline
KaLM-Embedding/KaLM-embed & STS15 & Retrieve semantically similar text \\
\hline
KaLM-Embedding/KaLM-embed & STS14 & Retrieve semantically similar text \\
\hline
KaLM-Embedding/KaLM-embed & MIRACLRetrievalHard & Given a question, retrieve Wikipedia passages that answer the question \\
\hline
KaLM-Embedding/KaLM-embed & TweetSentimentClassificat & Classify user passages. \\
\hline
KaLM-Embedding/KaLM-embed & MedrxivClusteringP2P.v2 & Identify the main category of Medrxiv papers based on the titles and abstracts \\
\hline
KaLM-Embedding/KaLM-embed & StackExchangeClustering.v & Identify the topic or theme of StackExchange posts based on the titles \\
\hline
KaLM-Embedding/KaLM-embed & ImdbClassification & Classify the sentiment expressed in the given movie review text from the IMDB dataset \\
\hline
\end{tabular}
\label{tab:default_model_task_prompts_2}
\end{table*}

\section{Best and Worst Performing Prompts} \label{app:best_worst_prompts}

\begin{table*}[t]
\centering
\small
\caption{Best and worst prompts for task: AmazCounterFact}
\begin{tabular}{|p{0.24\textwidth}|c|c|p{0.58\textwidth}|}
\toprule
Model & Type & Score & Prompt \\
\toprule
\multirow{2}{=}{\raggedright\texttt{KaLM-\allowbreak multiling-\allowbreak mini-\allowbreak 2.5}} & Best & 0.958 & \texttt{Category prediction for Amazon review counterfactual detection pair in eng: } \\
\midrule
& Worst & 0.808 & \texttt{Assign a category to this text:} \\
\multirow{2}{=}{\raggedright\texttt{Qwen3-\allowbreak 0.6B}} & Best & 0.931 & \texttt{Classify this Amazon review pair for counterfactual detection in eng: } \\
\midrule
& Worst & 0.67 & \texttt{Assign a classification label to these Amazon reviews in eng:} \\
\multirow{2}{=}{\raggedright\texttt{bge-\allowbreak base-\allowbreak en-\allowbreak 1.5}} & Best & 0.734 & \texttt{Label this example:} \\
\midrule
& Worst & 0.648 & \texttt{Category prediction for Amazon review counterfactual detection pair in eng: } \\
\multirow{2}{=}{\raggedright\texttt{bge-\allowbreak large-\allowbreak en-\allowbreak 1.5}} & Best & 0.737 & \texttt{Classify:} \\
\midrule
& Worst & 0.571 & \texttt{Counterfactual detection pair classification for Amazon customer reviews (eng):} \\
\multirow{2}{=}{\raggedright\texttt{bge-\allowbreak small-\allowbreak en-\allowbreak 1.5}} & Best & 0.711 & \texttt{Classify:} \\
\midrule
& Worst & 0.61 & \texttt{Counterfactual detection pair classification for Amazon customer reviews (eng):} \\
\multirow{2}{=}{\raggedright\texttt{multiling-\allowbreak e5-\allowbreak large}} & Best & 0.75 & \texttt{Assign a counterfactual detection label:} \\
\midrule
& Worst & 0.589 & \texttt{Predict the class:} \\
\bottomrule
\end{tabular}
\end{table*}

\begin{table*}[t]
\centering
\small
\caption{Best and worst prompts for task: FEVERHardNegatives}
\begin{tabular}{|p{0.24\textwidth}|c|c|p{0.58\textwidth}|}
\toprule
Model & Type & Score & Prompt \\
\toprule
\multirow{2}{=}{\raggedright\texttt{KaLM-\allowbreak multiling-\allowbreak mini-\allowbreak 2.5}} & Best & 0.882 & \texttt{default} \\
\midrule
& Worst & 0.79 & \texttt{Relevant passage retrieval in eng (FEVER dataset):} \\
\multirow{2}{=}{\raggedright\texttt{Qwen3-\allowbreak 0.6B}} & Best & 0.89 & \texttt{Retrieval task prompt:} \\
\midrule
& Worst & 0.853 & \texttt{Encode this claim for retrieval:} \\
\multirow{2}{=}{\raggedright\texttt{bge-\allowbreak base-\allowbreak en-\allowbreak 1.5}} & Best & 0.86 & \texttt{default} \\
\midrule
& Worst & 0.5 & \texttt{Text representation for retrieval:} \\
\multirow{2}{=}{\raggedright\texttt{bge-\allowbreak large-\allowbreak en-\allowbreak 1.5}} & Best & 0.873 & \texttt{default} \\
\midrule
& Worst & 0.309 & \texttt{Text representation for retrieval:} \\
\multirow{2}{=}{\raggedright\texttt{bge-\allowbreak small-\allowbreak en-\allowbreak 1.5}} & Best & 0.862 & \texttt{default} \\
\midrule
& Worst & 0.507 & \texttt{Text representation for retrieval:} \\
\multirow{2}{=}{\raggedright\texttt{multiling-\allowbreak e5-\allowbreak large}} & Best & 0.832 & \texttt{Retrieve relevant documents in eng for the FEVER claim:} \\
\midrule
& Worst & 0.131 & \texttt{Encode this text for retrieval:} \\
\bottomrule
\end{tabular}
\end{table*}

\begin{table*}[t]
\centering
\small
\caption{Best and worst prompts for task: ImdbClassification}
\begin{tabular}{|p{0.24\textwidth}|c|c|p{0.58\textwidth}|}
\toprule
Model & Type & Score & Prompt \\
\toprule
\multirow{2}{=}{\raggedright\texttt{KaLM-\allowbreak multiling-\allowbreak mini-\allowbreak 2.5}} & Best & 0.959 & \texttt{default} \\
\midrule
& Worst & 0.945 & \texttt{Determine the class for the given passage:} \\
\multirow{2}{=}{\raggedright\texttt{Qwen3-\allowbreak 0.6B}} & Best & 0.954 & \texttt{default} \\
\midrule
& Worst & 0.894 & \texttt{Determine the class for the given passage:} \\
\multirow{2}{=}{\raggedright\texttt{bge-\allowbreak base-\allowbreak en-\allowbreak 1.5}} & Best & 0.914 & \texttt{Assign a label to this review:} \\
\midrule
& Worst & 0.829 & \texttt{Passage: eng - Classify this example from the Large Movie Review Dataset:} \\
\multirow{2}{=}{\raggedright\texttt{bge-\allowbreak large-\allowbreak en-\allowbreak 1.5}} & Best & 0.935 & \texttt{Assign a label to this review:} \\
\midrule
& Worst & 0.798 & \texttt{Passage: eng - Classify this example from the Large Movie Review Dataset:} \\
\multirow{2}{=}{\raggedright\texttt{bge-\allowbreak small-\allowbreak en-\allowbreak 1.5}} & Best & 0.917 & \texttt{Assign a label to this review:} \\
\midrule
& Worst & 0.78 & \texttt{Passage: eng - Classify this example from the Large Movie Review Dataset:} \\
\multirow{2}{=}{\raggedright\texttt{multiling-\allowbreak e5-\allowbreak large}} & Best & 0.95 & \texttt{Classify the movie review:} \\
\midrule
& Worst & 0.75 & \texttt{Determine the class for the given passage:} \\
\bottomrule
\end{tabular}
\end{table*}

\begin{table*}[t]
\centering
\small
\caption{Best and worst prompts for task: MIRACLRetrievalHardNegatives.v2}
\begin{tabular}{|p{0.24\textwidth}|c|c|p{0.58\textwidth}|}
\toprule
Model & Type & Score & Prompt \\
\toprule
\multirow{2}{=}{\raggedright\texttt{KaLM-\allowbreak multiling-\allowbreak mini-\allowbreak 2.5}} & Best & 0.556 & \texttt{Search for documents using this query:} \\
\midrule
& Worst & 0.461 & \texttt{Instruction for MIRACL retrieval task:} \\
\multirow{2}{=}{\raggedright\texttt{Qwen3-\allowbreak 0.6B}} & Best & 0.686 & \texttt{default} \\
\midrule
& Worst & 0.485 & \texttt{Text to encode for multilingual retrieval in eng:} \\
\multirow{2}{=}{\raggedright\texttt{bge-\allowbreak base-\allowbreak en-\allowbreak 1.5}} & Best & 0.481 & \texttt{Search query:} \\
\midrule
& Worst & 0.076 & \texttt{Text to encode for multilingual retrieval in eng:} \\
\multirow{2}{=}{\raggedright\texttt{bge-\allowbreak large-\allowbreak en-\allowbreak 1.5}} & Best & 0.475 & \texttt{Search query:} \\
\midrule
& Worst & 0.048 & \texttt{Query representation for retrieval:} \\
\multirow{2}{=}{\raggedright\texttt{bge-\allowbreak small-\allowbreak en-\allowbreak 1.5}} & Best & 0.455 & \texttt{Search query:} \\
\midrule
& Worst & 0.019 & \texttt{Encode query for multilingual retrieval:} \\
\multirow{2}{=}{\raggedright\texttt{multiling-\allowbreak e5-\allowbreak large}} & Best & 0.621 & \texttt{default} \\
\midrule
& Worst & 0.089 & \texttt{Encode this text for retrieval:} \\
\bottomrule
\end{tabular}
\end{table*}

\begin{table*}[t]
\centering
\small
\caption{Best and worst prompts for task: MedrxivClusteringP2P.v2}
\begin{tabular}{|p{0.24\textwidth}|c|c|p{0.58\textwidth}|}
\toprule
Model & Type & Score & Prompt \\
\toprule
\multirow{2}{=}{\raggedright\texttt{KaLM-\allowbreak multiling-\allowbreak mini-\allowbreak 2.5}} & Best & 0.456 & \texttt{default} \\
\midrule
& Worst & 0.355 & \texttt{Group similar texts:} \\
\multirow{2}{=}{\raggedright\texttt{Qwen3-\allowbreak 0.6B}} & Best & 0.422 & \texttt{default} \\
\midrule
& Worst & 0.368 & \texttt{Medrxiv title+abstract clustering task:} \\
\multirow{2}{=}{\raggedright\texttt{bge-\allowbreak base-\allowbreak en-\allowbreak 1.5}} & Best & 0.371 & \texttt{Cluster representation:} \\
\midrule
& Worst & 0.362 & \texttt{Text clustering input:} \\
\multirow{2}{=}{\raggedright\texttt{bge-\allowbreak large-\allowbreak en-\allowbreak 1.5}} & Best & 0.367 & \texttt{Clustering input:} \\
\midrule
& Worst & 0.356 & \texttt{medrxiv titles+abstract clustering input:} \\
\multirow{2}{=}{\raggedright\texttt{bge-\allowbreak small-\allowbreak en-\allowbreak 1.5}} & Best & 0.367 & \texttt{default} \\
\midrule
& Worst & 0.359 & \texttt{Text clustering input:} \\
\multirow{2}{=}{\raggedright\texttt{multiling-\allowbreak e5-\allowbreak large}} & Best & 0.399 & \texttt{Clustering of titles+abstract from medrxiv across 51 categories:} \\
\midrule
& Worst & 0.356 & \texttt{Group this text:} \\
\bottomrule
\end{tabular}
\end{table*}

\begin{table*}[t]
\centering
\small
\caption{Best and worst prompts for task: STS14}
\begin{tabular}{|p{0.24\textwidth}|c|c|p{0.58\textwidth}|}
\toprule
Model & Type & Score & Prompt \\
\toprule
\multirow{2}{=}{\raggedright\texttt{KaLM-\allowbreak multiling-\allowbreak mini-\allowbreak 2.5}} & Best & 0.86 & \texttt{default} \\
\midrule
& Worst & 0.713 & \texttt{Measure semantic similarity for eng inputs from SemEval STS 2014 dataset:} \\
\multirow{2}{=}{\raggedright\texttt{Qwen3-\allowbreak 0.6B}} & Best & 0.871 & \texttt{default} \\
\midrule
& Worst & 0.692 & \texttt{Text pair similarity assessment (SemEval STS 2014, eng): } \\
\multirow{2}{=}{\raggedright\texttt{bge-\allowbreak base-\allowbreak en-\allowbreak 1.5}} & Best & 0.823 & \texttt{default} \\
\midrule
& Worst & 0.597 & \texttt{Text pair similarity assessment (SemEval STS 2014, eng): } \\
\multirow{2}{=}{\raggedright\texttt{bge-\allowbreak large-\allowbreak en-\allowbreak 1.5}} & Best & 0.828 & \texttt{default} \\
\midrule
& Worst & 0.582 & \texttt{Text pair similarity assessment:} \\
\multirow{2}{=}{\raggedright\texttt{bge-\allowbreak small-\allowbreak en-\allowbreak 1.5}} & Best & 0.818 & \texttt{default} \\
\midrule
& Worst & 0.562 & \texttt{Text pair similarity assessment:} \\
\multirow{2}{=}{\raggedright\texttt{multiling-\allowbreak e5-\allowbreak large}} & Best & 0.848 & \texttt{default} \\
\midrule
& Worst & 0.781 & \texttt{Measure semantic similarity between:} \\
\bottomrule
\end{tabular}
\end{table*}

\begin{table*}[t]
\centering
\small
\caption{Best and worst prompts for task: STS15}
\begin{tabular}{|p{0.24\textwidth}|c|c|p{0.58\textwidth}|}
\toprule
Model & Type & Score & Prompt \\
\toprule
\multirow{2}{=}{\raggedright\texttt{KaLM-\allowbreak multiling-\allowbreak mini-\allowbreak 2.5}} & Best & 0.903 & \texttt{default} \\
\midrule
& Worst & 0.821 & \texttt{Text similarity level in SemEval STS 2015 dataset:} \\
\multirow{2}{=}{\raggedright\texttt{Qwen3-\allowbreak 0.6B}} & Best & 0.914 & \texttt{default} \\
\midrule
& Worst & 0.784 & \texttt{SemEval STS 2015 dataset - measure similarity between pairs: } \\
\multirow{2}{=}{\raggedright\texttt{bge-\allowbreak base-\allowbreak en-\allowbreak 1.5}} & Best & 0.88 & \texttt{default} \\
\midrule
& Worst & 0.7 & \texttt{Measure semantic similarity in eng based on SemEval STS 2015 dataset:} \\
\multirow{2}{=}{\raggedright\texttt{bge-\allowbreak large-\allowbreak en-\allowbreak 1.5}} & Best & 0.88 & \texttt{default} \\
\midrule
& Worst & 0.693 & \texttt{STS 2015 style semantic similarity evaluation: } \\
\multirow{2}{=}{\raggedright\texttt{bge-\allowbreak small-\allowbreak en-\allowbreak 1.5}} & Best & 0.873 & \texttt{default} \\
\midrule
& Worst & 0.674 & \texttt{Measure text similarity in eng, SemEval STS 2015:} \\
\multirow{2}{=}{\raggedright\texttt{multiling-\allowbreak e5-\allowbreak large}} & Best & 0.91 & \texttt{default} \\
\midrule
& Worst & 0.855 & \texttt{Measure semantic similarity between:} \\
\bottomrule
\end{tabular}
\end{table*}

\begin{table*}[t]
\centering
\small
\caption{Best and worst prompts for task: STS22.v2}
\begin{tabular}{|p{0.24\textwidth}|c|c|p{0.58\textwidth}|}
\toprule
Model & Type & Score & Prompt \\
\toprule
\multirow{2}{=}{\raggedright\texttt{KaLM-\allowbreak multiling-\allowbreak mini-\allowbreak 2.5}} & Best & 0.732 & \texttt{default} \\
\midrule
& Worst & 0.685 & \texttt{Compare texts for semantic similarity:} \\
\multirow{2}{=}{\raggedright\texttt{Qwen3-\allowbreak 0.6B}} & Best & 0.718 & \texttt{default} \\
\midrule
& Worst & 0.646 & \texttt{Query: assess text pair similarity} \\
\multirow{2}{=}{\raggedright\texttt{bge-\allowbreak base-\allowbreak en-\allowbreak 1.5}} & Best & 0.678 & \texttt{In eng, assess similarity of news article pair – SemEval 2022 Task 8:} \\
\midrule
& Worst & 0.466 & \texttt{default} \\
\multirow{2}{=}{\raggedright\texttt{bge-\allowbreak large-\allowbreak en-\allowbreak 1.5}} & Best & 0.699 & \texttt{Similarity between texts in eng:} \\
\midrule
& Worst & 0.489 & \texttt{default} \\
\multirow{2}{=}{\raggedright\texttt{bge-\allowbreak small-\allowbreak en-\allowbreak 1.5}} & Best & 0.674 & \texttt{Compare the similarity of these SemEval 2022 Task 8 news articles:} \\
\midrule
& Worst & 0.469 & \texttt{default} \\
\multirow{2}{=}{\raggedright\texttt{multiling-\allowbreak e5-\allowbreak large}} & Best & 0.69 & \texttt{default} \\
\midrule
& Worst & 0.638 & \texttt{Similarity score for article pair:} \\
\bottomrule
\end{tabular}
\end{table*}

\begin{table*}[t]
\centering
\small
\caption{Best and worst prompts for task: StackExchangeClustering.v2}
\begin{tabular}{|p{0.24\textwidth}|c|c|p{0.58\textwidth}|}
\toprule
Model & Type & Score & Prompt \\
\toprule
\multirow{2}{=}{\raggedright\texttt{KaLM-\allowbreak multiling-\allowbreak mini-\allowbreak 2.5}} & Best & 0.777 & \texttt{Cluster these StackExchange titles:} \\
\midrule
& Worst & 0.556 & \texttt{Group similar texts:} \\
\multirow{2}{=}{\raggedright\texttt{Qwen3-\allowbreak 0.6B}} & Best & 0.712 & \texttt{default} \\
\midrule
& Worst & 0.553 & \texttt{eng text for clustering of 25 sets, each with 10-50 classes and 100-1000 sentences per class:} \\
\multirow{2}{=}{\raggedright\texttt{bge-\allowbreak base-\allowbreak en-\allowbreak 1.5}} & Best & 0.58 & \texttt{Grouping for clustering:} \\
\midrule
& Worst & 0.486 & \texttt{Encode for clustering: titles from 121 stackexchanges in eng} \\
\multirow{2}{=}{\raggedright\texttt{bge-\allowbreak large-\allowbreak en-\allowbreak 1.5}} & Best & 0.594 & \texttt{default} \\
\midrule
& Worst & 0.5 & \texttt{Encode for clustering: titles from 121 stackexchanges in eng} \\
\multirow{2}{=}{\raggedright\texttt{bge-\allowbreak small-\allowbreak en-\allowbreak 1.5}} & Best & 0.573 & \texttt{default} \\
\midrule
& Worst & 0.481 & \texttt{Group similar texts into clusters: Clustering of titles from 121 stackexchanges in eng} \\
\multirow{2}{=}{\raggedright\texttt{multiling-\allowbreak e5-\allowbreak large}} & Best & 0.672 & \texttt{Cluster these StackExchange titles:} \\
\midrule
& Worst & 0.6 & \texttt{default} \\
\bottomrule
\end{tabular}
\end{table*}

\begin{table*}[t]
\centering
\small
\caption{Best and worst prompts for task: Touche2020Retrieval.v3}
\begin{tabular}{|p{0.24\textwidth}|c|c|p{0.58\textwidth}|}
\toprule
Model & Type & Score & Prompt \\
\toprule
\multirow{2}{=}{\raggedright\texttt{KaLM-\allowbreak multiling-\allowbreak mini-\allowbreak 2.5}} & Best & 0.644 & \texttt{Retrieval query:} \\
\midrule
& Worst & 0.218 & \texttt{Representation for retrieval in eng, Touché Task 1: Argument Retrieval for Controversial Questions} \\
\multirow{2}{=}{\raggedright\texttt{Qwen3-\allowbreak 0.6B}} & Best & 0.699 & \texttt{default} \\
\midrule
& Worst & 0.383 & \texttt{Encode (eng) for Touché Task 1: Argument Retrieval for Controversial Questions:} \\
\multirow{2}{=}{\raggedright\texttt{bge-\allowbreak base-\allowbreak en-\allowbreak 1.5}} & Best & 0.497 & \texttt{default} \\
\midrule
& Worst & 0.035 & \texttt{Representation for retrieval in eng, Touché Task 1: Argument Retrieval for Controversial Questions} \\
\multirow{2}{=}{\raggedright\texttt{bge-\allowbreak large-\allowbreak en-\allowbreak 1.5}} & Best & 0.471 & \texttt{default} \\
\midrule
& Worst & 0.014 & \texttt{Retrieve relevant argument in eng for Touché Task 1: Argument Retrieval for Controversial Questions:} \\
\multirow{2}{=}{\raggedright\texttt{bge-\allowbreak small-\allowbreak en-\allowbreak 1.5}} & Best & 0.576 & \texttt{default} \\
\midrule
& Worst & 0.009 & \texttt{Retrieve relevant argument in eng for Touché Task 1: Argument Retrieval for Controversial Questions:} \\
\multirow{2}{=}{\raggedright\texttt{multiling-\allowbreak e5-\allowbreak large}} & Best & 0.545 & \texttt{Search query:} \\
\midrule
& Worst & 0.342 & \texttt{Encode for Touché Task 1 retrieval:} \\
\bottomrule
\end{tabular}
\end{table*}

\begin{table*}[t]
\centering
\small
\caption{Best and worst prompts for task: TweetSentimentClassification}
\begin{tabular}{|p{0.24\textwidth}|c|c|p{0.58\textwidth}|}
\toprule
Model & Type & Score & Prompt \\
\toprule
\multirow{2}{=}{\raggedright\texttt{KaLM-\allowbreak multiling-\allowbreak mini-\allowbreak 2.5}} & Best & 0.704 & \texttt{Classify the sentiment of this tweet:} \\
\midrule
& Worst & 0.526 & \texttt{Category prediction:} \\
\multirow{2}{=}{\raggedright\texttt{Qwen3-\allowbreak 0.6B}} & Best & 0.706 & \texttt{Classify the sentiment of this tweet:} \\
\midrule
& Worst & 0.483 & \texttt{Category prediction:} \\
\multirow{2}{=}{\raggedright\texttt{bge-\allowbreak base-\allowbreak en-\allowbreak 1.5}} & Best & 0.518 & \texttt{Label this example:} \\
\midrule
& Worst & 0.417 & \texttt{default} \\
\multirow{2}{=}{\raggedright\texttt{bge-\allowbreak large-\allowbreak en-\allowbreak 1.5}} & Best & 0.529 & \texttt{Label the following:} \\
\midrule
& Worst & 0.427 & \texttt{default} \\
\multirow{2}{=}{\raggedright\texttt{bge-\allowbreak small-\allowbreak en-\allowbreak 1.5}} & Best & 0.52 & \texttt{Classify this tweet in eng:} \\
\midrule
& Worst & 0.417 & \texttt{default} \\
\multirow{2}{=}{\raggedright\texttt{multiling-\allowbreak e5-\allowbreak large}} & Best & 0.645 & \texttt{eng sentiment analysis – classify:} \\
\midrule
& Worst & 0.485 & \texttt{Determine the class for the given passage:} \\
\bottomrule
\end{tabular}
\end{table*}

\section{AI Tools Usage}
In this work, we utilized AI tools to assist with code generation, debugging, and spell-checking/grammatical editing of the manuscript text. Specifically, we used Anthropic’s Claude models \cite{anthropic_claude} and OpenAI’s ChatGPT \cite{singh2025openaigpt5card}.

\end{document}